\documentclass[twoside,leqno,twocolumn]{article}

\usepackage[letterpaper]{geometry}

\usepackage{ltexpprt}
\usepackage{hyperref}
\usepackage{cite}
\usepackage{amsmath,amssymb,amsfonts}
\usepackage{algorithmic}
\usepackage{graphicx}
\usepackage{textcomp}
\usepackage{xcolor}
\usepackage{bbold}
\usepackage{hyperref}
\usepackage{multirow}
\usepackage{booktabs}
\usepackage{amsmath}

\DeclareMathOperator*{\argmin}{arg\,min}

\begin{document}

\newcommand\relatedversion{}
\renewcommand\relatedversion{\thanks{The full version of the paper can be accessed at \protect\url{https://arxiv.org/abs/1902.09310}}} 

\title{\Large Manifold Alignment with Label Information}
\author{Andr\'es F. Duque\thanks{Utah State University, andres.duque@usu.edu}
\and Myriam Lizotte\thanks{Universit\'e de Montr\'eal, myriam.lizotte@mila.quebec }
\and Guy Wolf \thanks{Universit\'e de Montr\'eal, wolfguy@mila.quebec}
\and Kevin R. Moon \thanks{Utah State University, kevin.moon@usu.edu}
}

\date{}

\maketitle







\begin{abstract} \small\baselineskip=9pt Multi-domain data is becoming increasingly common and presents both challenges and opportunities in the data science community. The integration of distinct data-views can be used for exploratory data analysis, and benefit downstream analysis including machine learning related tasks. With this in mind, we present a novel manifold alignment method called MALI (Manifold alignment with label information) that learns a correspondence between two distinct domains.  MALI can be considered as belonging to a middle ground between the more commonly addressed \textit{semi-supervised} manifold alignment problem with some known correspondences between the two domains,  and the purely \textit{unsupervised} case, where no known correspondences are provided. To do this, MALI learns the manifold structure in both domains via a diffusion process and then leverages discrete class labels to guide the alignment. By aligning two distinct domains, MALI  recovers a pairing and a common representation that reveals related samples in both domains. Additionally, MALI can be used for the transfer learning problem known as domain adaptation. We show that MALI outperforms the current state-of-the-art manifold alignment methods across multiple datasets. 

\end{abstract}

\section{Introduction}

The data collection process of a given phenomena may be affected by different sources of variability, creating seemingly distinct domains. For instance, natural images with different illumination, contrast or noise, may affect the classification performance of a machine learning model previously trained on a different domain. In biology, the modern study of single-cell dynamics is conducted via different instruments, conditions and modalities, raising different challenges and opportunities \cite{stuart2019integrative, stuart2019comprehensive}. In many cases, the relationships between the different domains are unknown. Hence, the fusion and integration of multi-domain data has been extensively studied in the data science community for supervised learning as well as data mining and exploratory data analysis. One of the earliest methods to do this is Canonical Correlation Analysis (CCA), which finds a linear projection that maximizes the correlation between the two domains~\cite{thompson1984canonical}. CCA has been extended to different formulations in recent years as sparse  CCA~\cite{hardoon2011sparse, parkhomenko2009sparse} or Kernel CCA~\cite{gao2012multi,chang2013canonical}. 

\begin{figure}
\centering
    \includegraphics[width =  \columnwidth]{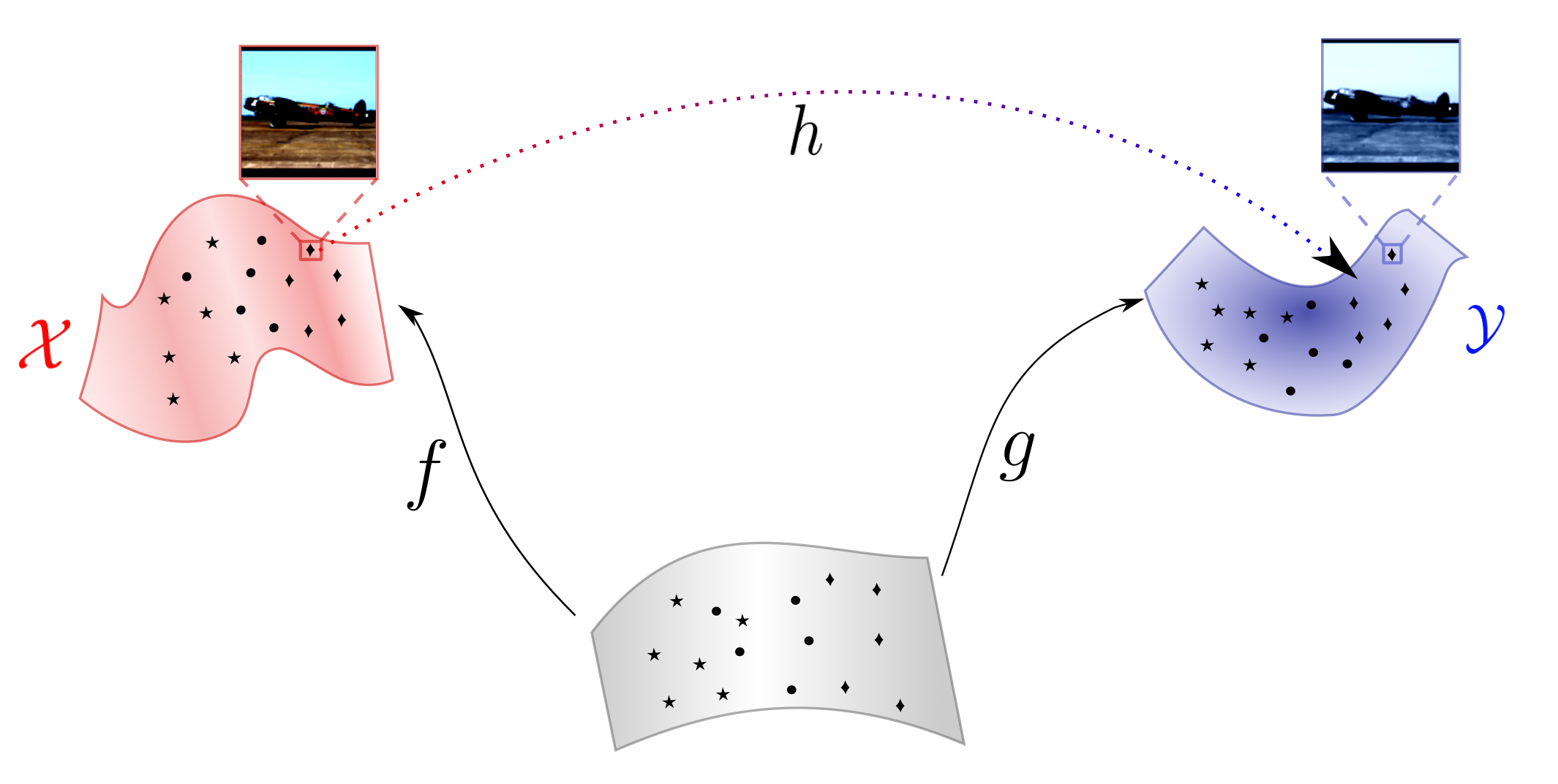}
    \caption{\textbf{Manifold alignment}. Two different datasets measured from the same underlying phenomena are captured in different conditions, instruments, experimental designs, etc. Manifold alignment assumes a common latent space (grey) from which the observations are mapped by functions $f$ and $g$ to the different ambient spaces. We seek to find the underlying relationship $h$ between observations living in different spaces $\mathcal{X}$ and $\mathcal{Y}$ without assuming any pairing known \textit{a priori}. Instead we assume there are labeled observations for different classes (different shapes).} 
    \label{fig:manf_align_concept}
\end{figure}

In many applications, a reasonable assumption to make is that the data collected in different domains is controlled by a set of shared underlying modes of variation or latent variables. The manifold assumption is also often applicable in this case, in which  the data measured in the different domains are assumed to lie on a low-dimensional manifold embedded in the high-dimensional ambient spaces, being the result of smooth mappings of the latent variables (see Fig.~\ref{fig:manf_align_concept}). With this in mind, manifold alignment (MA) has become a common technique for data integration. Some applications of MA include handling different face poses and protein structure alignment (\cite{zhai122010manifold, abeo2019manifold}),  medical images for Alzheimer’s disease classification (\cite{baumgartner2015self}, \cite{guerrero2014manifold}), multi-modal sensing images \cite{tuia2014semisupervised}, graph-matching \cite{escolano2011graph}, and integrating single-cell multi-omics data \cite{cao2020unsupervised} . 

Multiple MA methods have been proposed under different prior knowledge assumptions that relate the two domains. Methods such as CCA or multi-view diffusion maps \cite{lindenbaum2020multi} can be categorized as \textit{supervised} MA, since the data is assumed to come in a paired fashion.  More challenging  scenarios arise when partial or null \textit{a priori} pairing knowledge is considered. 
Purely \textit{unsupervised} algorithms are designed for scenarios where neither pairings between domains nor any other side-information is available. As a consequence, they rely solely on the particular topology of each  domain to infer inter-domain similarities (e.g. \cite{wang2009manifold,cui2014generalized,cao2020unsupervised,demetci2022scot}). 

Methods that leverage some additional information are often categorized as \textit{semi-supervised} MA. As a special case, several methods consider partial correspondence information, where a few one-to-one matching samples work as anchor points to find a consistent alignment for the rest of the data. Some papers  leverage the graph structure of the data \cite{ham2003learning, ham2005semisupervised,wang2008manifold,duque2022diffusion} and are closely related to Laplacian eigenmaps \cite{belkin2003laplacian}. Others resort to neural networks such as the GAN-based  MAGAN \cite{amodio2018magan} or the autoencoder presented in \cite{aziz2019semi}. 

However, even partial correspondences can be expensive or impossible to acquire. This is the case in biological applications where the measurement process destroys the cells, making it impossible to measure other modalities of the exact same cells. But even if there are no known correspondences between domains, we do not have to resort to unsupervised MA. If we have access to side information about the datasets from both domains, such as discrete class labels, we can leverage this extra knowledge to perform manifold alignment~\cite{wang2011heterogeneous,tuia2016kernel,wang2019label}. Motivated by this, we propose a new semi-supervised MA algorithm called MALI (Manifold Alignment with Label Information). MALI  leverages the manifold structure of the data in both domains, combined with the discrete label information, and it does not require any known corresponding points in the different domains. MALI is built upon the widely-used manifold learning method Diffusion Maps \cite{coifman2006diffusion} and optimal transport (OT) \cite{peyre2019computational}. We show via experimentation that MALI outperforms  current state-of-the-art MA algorithms in this setting across multiple datasets by several metrics.

The  setting described above is similar to the domain adaptation (DA) \cite{domainadapt} problem.  In traditional machine learning, the training set and the test set are assumed to be sampled from the same distribution and to share the same features. But in practice these assumptions may not hold, for example, due to the different collection circumstances mentioned previously. When data is expensive or time-consuming to label, it may be desirable to train a model on existing related datasets and then adapt it to the new task. It is of interest to leverage the knowledge acquired from training on one dataset to improve performance on the same task on a different dataset, or potentially even a different task. One possible approach to tackle DA is to use MA, since knowledge can be transferred through MA via the learned inter-domain correspondences or by training on a shared latent representation of both domains.

\section{Preliminaries}

\subsection{Problem Description}

Assume we have two datasets $\mathcal{X}= \{x_{1}, x_{2}, \ldots,  x_{n}\} \in \mathrm{R}^{n \times p}$ and $\mathcal{Y} = \{y_{1}, y_{2}, \ldots,  y_{m}\} \in \mathrm{R}^{m \times q}$. We assume that all of the points in $\mathcal{X}$ are labeled with discrete (i.e. class) labels $\mathcal{L}_x=\{\ell_1^x,\dots,\ell_n^x \}$ while the points in $\mathcal{Y}$ may be partially or fully labeled with discrete labels $\mathcal{L}_y=\{\ell_1^y,\dots,\ell_r^y\}$, with $r\leq m$. In the domain adaptation problem, $\mathcal{X}$ is measured from the source domain while $\mathcal{Y}$ is measured from the target domain. 

The problem consists of learning an alignment between both data manifolds, by leveraging their respective geometric structures as well as the label knowledge available from both domains. There are several possible ways to represent such an alignment using MA algorithms. One way is to directly learn hard or soft correspondences between points in $\mathcal{X}$ and $\mathcal{Y}$. A regression model could then be trained using these correspondences to learn a parametric mapping between domains. In the domain adaptation problem, unlabeled data in the target domain can be labeled using the more rich label information using the regression model. 

A second way to represent the alignment is to learn a shared embedding space which can be used for downstream analysis. For the domain adaptation problem, a classifier could be trained on the shared embedding space using the labels from $\mathcal{X}$.  Direct correspondences can be learned by, for example, using a nearest neighbor approach in the shared space. 

As we show in this work, MALI is suited for any of these scenarios. We first find pairwise cross-domain distances, which are then leveraged to find hard or soft assignments between the domains via optimal transport. If required, a shared embedding can be learned using these assignments. 



\subsection{Related work}
\label{subsec:relatW}

Here we summarize two existing methods that perform manifold alignment using discrete label information without assuming prior known correspondences.  In \cite{wang2011heterogeneous} both datasets are concatenated in a new block matrix 
\begin{align*}
Z = \left[\begin{array}{cc}
\mathcal{X} & \boldsymbol{0}\\
\boldsymbol{0} & \mathcal{Y} \\
\end{array}\right] \in \mathbb{R}^{(n+m)\times (p+q)}.
\end{align*}
Domain-specific similarity matrices $W_{\mathcal{X}}$ and $W_{\mathcal{Y}}$ are created from the data, e.g. via a kernel function. These matrices are then  similarly combined in a new block matrix $W_{Z}$. To leverage the label information, the authors create a label-similarity matrix with entries $W_{s}(i,j) = 1$ if samples $z_{i}$ and $z_{j}$ share the same label, 0 otherwise. A label dissimilarity matrix $W_{d}(i,j) = |W_{s}(i,j) - 1|$ is also constructed.  The idea is to combine $W_{Z}$, which characterizes the topology of each domain, with the label-similarity matrix in the alignment. Finding a new representation where both domains are aligned, requires two conditions: 1) observations close to each other in their domains should remain close in the new representation; 2) if $z_{i}$ and $z_{j}$ belong to the same class they should be mapped close to each other, otherwise they should be distant. These conditions can be formally described using the graph Laplacian associated to each of the similarity/dissimilarity matrices, leading to an optimization problem with solution given by the generalized eigenvalue problem:
\begin{align}
    Z^{T}(\mu L_{z} + L_{s})Zv = \lambda Z^{T}L_{d}Zv, \label{eq:Eig}
\end{align}
where $L_z$, $L_s$, and $L_d$ are the graph Laplacian matrices of $W_z$, $W_s$, and $W_d$, respectively. For example, $L_z=D_z-W_z$, where $D_z$ is a diagonal matrix with diagonal entries corresponding to the sum of each row in $W_z$.

This formulation is a linear problem and thus cannot handle strong non-linear deformations \cite{tuia2016kernel}. Motivated by this shortcoming the authors of \cite{tuia2016kernel} extended the previous approach to a kernelized version called KEMA, with a new eigenvalue problem:
\begin{align}
    K(\mu L_{z} + L_{s})K\alpha = \lambda KL_{d}K\alpha, \label{eq:KemaEig}
\end{align}
with $K$ being a user-defined kernel matrix.
The aligned representation in both methods is obtained from the projection of the data onto the eigenvectors associated with the lowest non-zero eigenvalues in the generalized eigenvalue problems  \eqref{eq:Eig} and \eqref{eq:KemaEig}. That is, $U = Zv$ or $U = K\alpha$, respectively, for the first dimension.


\section{Manifold alignment with label information (MALI)}

The main idea behind MALI resides in finding an inter-domain distance or similarity between $x_{i}$ and $y_{j}$ that is leveraged to recover cross-domain relations. We start by building a graph from the data on each domain, where the weights of the edges connecting the nodes are computed via an $\alpha$-decay kernel~\cite{moon2019visualizing}:
\begin{align}
    W_{\mathcal{X}}(x_i,x_j)=&\frac{1}{2}\exp\left( -\frac{||x_i-x_j||^\alpha}{\sigma_{k}^\alpha(x_i)}\right) \nonumber \\
    &+  \frac{1}{2}\exp\left( -\frac{||x_i-x_j||^\alpha}{\sigma_{k}^\alpha(x_j)}\right),
    \label{eq:DecayGaussKern}
\end{align}
where $\alpha>0$ and $\sigma_k(x_i)$ is the $k$-nearest neighbor distance of $x_i$ in $\mathcal{X}$. The kernel $W_\mathcal{Y}$ is constructed similarly. The hyperparameters $\alpha$ and $k$ control the connectivity and local geometry preservation in the graph, although most methods are typically robust to these hyperparameters when using this kernel~\cite{moon2019visualizing}. For the  experiments presented in this work we set $\alpha = 10$ and $k= 10$.

 By row-normalizing $W_{*}$ ($* \in\{\mathcal{X}, \mathcal{Y}\}$) using their corresponding degree matrices $D_{*}$, we obtain the respective diffusion operators  $P_{\mathcal{X}} =  D^{-1}_{\mathcal{X}}W_{\mathcal{X}} \in \mathrm{R}^{n \times n}$ and $P_{\mathcal{Y}} = D^{-1}_{\mathcal{Y}}W_{\mathcal{Y}} \in \mathrm{R}^{m \times m}$. The diffusion operator  represents a probability transition matrix for the graph represented with the kernel matrix $W_*$. Typically, the diffusion operator is exponentiated $P_{*}^{t}$  to describe the transition probabilities between observations after $t$ steps in a random walk~\cite{coifman2006diffusion,moon2019visualizing}. Instead, we build a time-independent similarity matrix by aggregating the transition probabilities from every possible $t$-step random walk as follows:
 \begin{align}
    M_{*} = \sum_{t = 1}^{\infty} (P_{*} - \mathbb{1}\phi_{0}^{T})^{t} = (I - (P_{*} - \mathbb{1}\phi_{0}^{T}))^{-1} - I,
\end{align}
where $\mathbb{1}$ is a vector of ones and $\phi_0$ is the stationary distribution of the Markov chain, represented by the first left eigenvector of $P_*$. Subtracting $\mathbb{1}\phi_{0}^{T}$ from $P_*$ enables the series to converge.

 This construction was previously developed in \cite{haghverdi2016diffusion}, and constitutes the key quantity for the calculation of diffusion pseudotime (DPT). The advantages of working with $M$ instead of $P^{t}$ are twofold. First, we do not need to select the hyper-parameter $t$, which will be dataset dependent.  Given the nature of the problem, a cross-validation-based approach for hyper-parameter tuning is not possible. Second, the matrix $M$ builds a more extensive connection across the data as it considers the relationships between points at different scales of random walks. This does have the effect of smoothing the local relationships and the fine granular similarities are lost. However, this is not a problem as MALI focuses more on  coarse similarities as we describe next. 

To find inter-domain similarities, we need a new feature representation shared by the two domains. For this purpose, we use the label information, since it is the only available cross-domain information we possess a priori. Therefore, we aggregate the similarities of each observation grouped by each of the labels as follows:
\begin{align}
M_{*}^{l}(i, c) = \frac{1}{p^{*}_{c}}\sum_{j \in \mathcal{I}^{*}_{c}}M_{*}(i, j),    
\end{align}
where $\mathcal{I}^{*}_{c}$ is the set of indices in domain $*$ labeled with class $c \in \mathcal{C}$, with $\mathcal{C}$ denoting the set of all labels present in both domains, and $p^*_c$ is the estimated prior class probability (e.g. $p^\mathcal{X}_c=\frac{1}{n}\sum_{i=1}^n \mathbf{1}\{\ell_i^x=c\}$ where $\mathbf{1}\{\cdot\}$ is the indicator function). Normalizing by the priors $p^{*}_{c}$ accounts for class unbalance. The matrix $M_{*}^{l} \in \mathrm{R}^{\cdot \times |\mathcal{C}|}$ encodes a coarse similarity between samples and labels. Notice that we now have a  similarity between $\mathcal{X}$ and $\mathcal{Y}$ and the labels, even though the datasets may have come from different spaces with different dimensionality. This allows us to compute inter-domain cosine distances between points $x_i\in \mathcal{X}$ and $y_j\in\mathcal{Y}$:  
\begin{equation}
    \begin{split}
    D_{ij} = \left(1 - \frac{\langle M^{l}_{\mathcal{X}}(i,:) , M^{l}_{\mathcal{Y}}(j, :)\rangle}{||M^{l}_{\mathcal{X}}(i,:)||||M^{l}_{\mathcal{Y}}(j,:)||}\right). 
    \label{eq:interDistances}
    \end{split}
\end{equation}

Matrix $D$  contains the information we need  for recovering a matching among samples from both domains and for learning a common representation. One way to find a matching is to construct a matrix $T$ with entries $T_{ij} = 1$ if $y_{j}$ is the nearest neighbor of $x_{i}$ among all observations in $\mathcal{Y}$ with respect to the distances given by \eqref{eq:interDistances}. Another  reasonable alternative is to construct a soft assignment using  $k$-nearest neighbors. However, in our experimental results we found improved performance by solving an entropic optimal transport problem instead \cite{peyre2019computational}, in which $D$ is the cost matrix: 
\begin{align}
T = \argmin_{T \in \mathcal{U}} \langle T, D \rangle_{F} + \epsilon\Omega(T).
\label{eq:regularization}
\end{align}

$T$ is sometimes referred to as a coupling matrix and belongs to the set $\mathcal{U}(\boldsymbol{a}, \boldsymbol{b}) = \{T \in \mathbb{R}^{n \times m}_{+}: T\mathbb{1}_{m} = \boldsymbol{a}, T^{T}\mathbb{1}_{n} = \boldsymbol{b}\}$, where $\boldsymbol{a}$ and $\boldsymbol{b}$ are vectors containing a user-defined ``mass'' for each observation. Matrix $T$ is the primary quantity of interest and  output of MALI, as it contains the coupling information between all the samples. The entropic regularization imposed by $\Omega(T) = \sum_{ij}T_{ij}\text{log}(T_{ij})$ drives the solution towards soft assignments. 

The matrix $T$ can be used to  project the data into a common representation. One approach to do this is to project directly into one of the ambient spaces using the barycentric projection, e.g. $x_{i} \mapsto \sum_{j} T_{ij}y_{j}$. This mapping is represented in Figure \ref{fig:manf_align_concept} by $h$. Alternatively, a joint latent representation can be obtained after computing a spectral embedding \cite{belkin2003laplacian}  on the cross-domain similarity:   
\begin{align}
\label{eq:w_matrix}
W = 
\left[\begin{array}{cc}
 \mu W_{\mathcal{X}} & (1-\mu) W_{\mathcal{X}\mathcal{Y}}\\  
 (1-\mu) W^{T}_{\mathcal{X}\mathcal{Y}} &  \mu W_{\mathcal{Y}} 
\end{array}
\right],
\end{align}
 where $\mu$ controls the preservation of the domain specific topology, and the off-diagonal blocks in $W$ reproduce the inter-domain similarities according to the assignments in $T$; i.e.  $W_{\mathcal{X}\mathcal{Y}} = (W_{\mathcal{X}}T + TW_{\mathcal{Y}})$. In Section~\ref{sub:dimension}, we empirically demonstrate that this construction of $W$ is more beneficial than including only the $T$ matrix in the off-diagonal blocks, as used in \cite{ham2005semisupervised}.

MALI differs from KEMA and the approach in~\cite{wang2011heterogeneous}, which directly find a latent joint representation. In contrast, MALI finds a coupling first via the distances in (\ref{eq:interDistances}), and then builds a joint similarity matrix, which if needed, can be used to find a latent joint representation.

\section{Experimental Results}

\subsection{Experimental setting}
\label{sec:results}
\begin{figure}
\centering
    \includegraphics[width=\columnwidth]{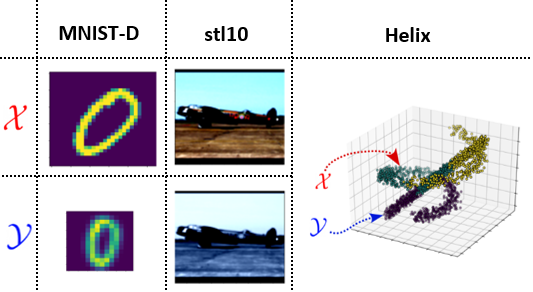}
    \caption{\textbf{Experimental datasets}. Three of the employed datasets highlighting the differences between the domains $\mathcal{X}$ and $\mathcal{Y}$. For the MNIST-D dataset, $\mathcal{X}$ contains the original images while $\mathcal{Y}$ contains transformed versions of the images. This also applies to the stl10 dataset. The helix dataset contains a helix ($\mathcal{X}$) and a straight line ($\mathcal{Y}$).}
    \label{fig:grid_domains}
\end{figure}

We now compare the performance of MALI with KEMArbf \cite{tuia2016kernel} and \cite{wang2011heterogeneous}, which we refer to as KEMAlin since it produces similar results as those using KEMA with a linear kernel. We use the code provided by the authors\footnote{\url{https://github.com/dtuia/KEMA}}. We perform the comparisons for the four datasets described next and displayed in Fig.~\ref{fig:grid_domains}:

\begin{itemize}
    \item Helix: two one-dimensional manifolds embedded in a 3D space plus noise, where one domain is a helix and the other a straight line.
    \item MNIST-D: one domain consists of the original MNIST digits, while the other is generated by applying multiple transformations including rotation, downscaling, and Gaussian blurring (see Fig.~\ref{fig:grid_domains}).
    \item stl10: a popular dataset for computer vision, where  the first domain contains the original images, and the second is generated by applying brightness, gray scaling, and Gaussian blurring. We performed feature extraction using the 512 outputs after the last convolution layer in ResNet-18, a smaller version of the ResNet neural network~\cite{he2016deep}. 
    \item RNA-ATAC:  Jointly measured observations of single-cell RNA-seq and ATAC-seq data. The data is available at  \href{https://openproblems.bio/neurips_2021/}{\textit{Multimodal Single-Cell Data Integration}} challenge, NeurIPS competition track 2021. We selected  batches ``s1d1'' and preprocessed both domains reducing their dimensionality to 1000 features via truncated SVD. 

\end{itemize}

For all of these datasets, we have access to the correspondences between points, although this information is not provided to any of the algorithms. A good alignment should map  matching observations close to each other, as well as be useful for classifying unlabeled target observations employing the labeled source samples. The following two metrics meet these requirements, and have been previously employed in \cite{cao2020unsupervised}, \cite{cao2022manifold}, \cite{demetci2022scot}. 
\begin{enumerate}
    \item Fraction of samples closer to the true match (FOSCTTM): Given the ground truth one-to-one correspondence knowledge, this metric computes how many samples are closer to the true match after alignment,  normalized by the total size of the data. Then the average error for all the samples is computed, yielding a final score. A perfect alignment would produce a score equal to zero.    
    \item Label Transfer: Using the labels of the source domain, a 1-NN classifier is built after alignment and tested on the target domain. The final score corresponds to the percentage of correctly predicted labels on the target domain.
\end{enumerate}

For now, we restrict ourselves to the  case where $a_{i} = 1,  \forall x_{i} \in \mathcal{X}$, $b_{j} = 1, \forall y_{j} \in \mathcal{Y}$, and $\epsilon=0$. Since both domains are the same size $n=m$, and one-to-one correspondences exist in these datasets, the selected parameters force matrix $T$ to be a zero-one matrix. However,  we relax these assumptions in Sections~\ref{sub:soft} and \ref{sub:unbalance} to show alternatives scenarios.



\subsection{Selecting the dimension for the alignment}
\label{sub:dimension}
\begin{figure}[!h]
\centering
    \includegraphics[width = \columnwidth ] {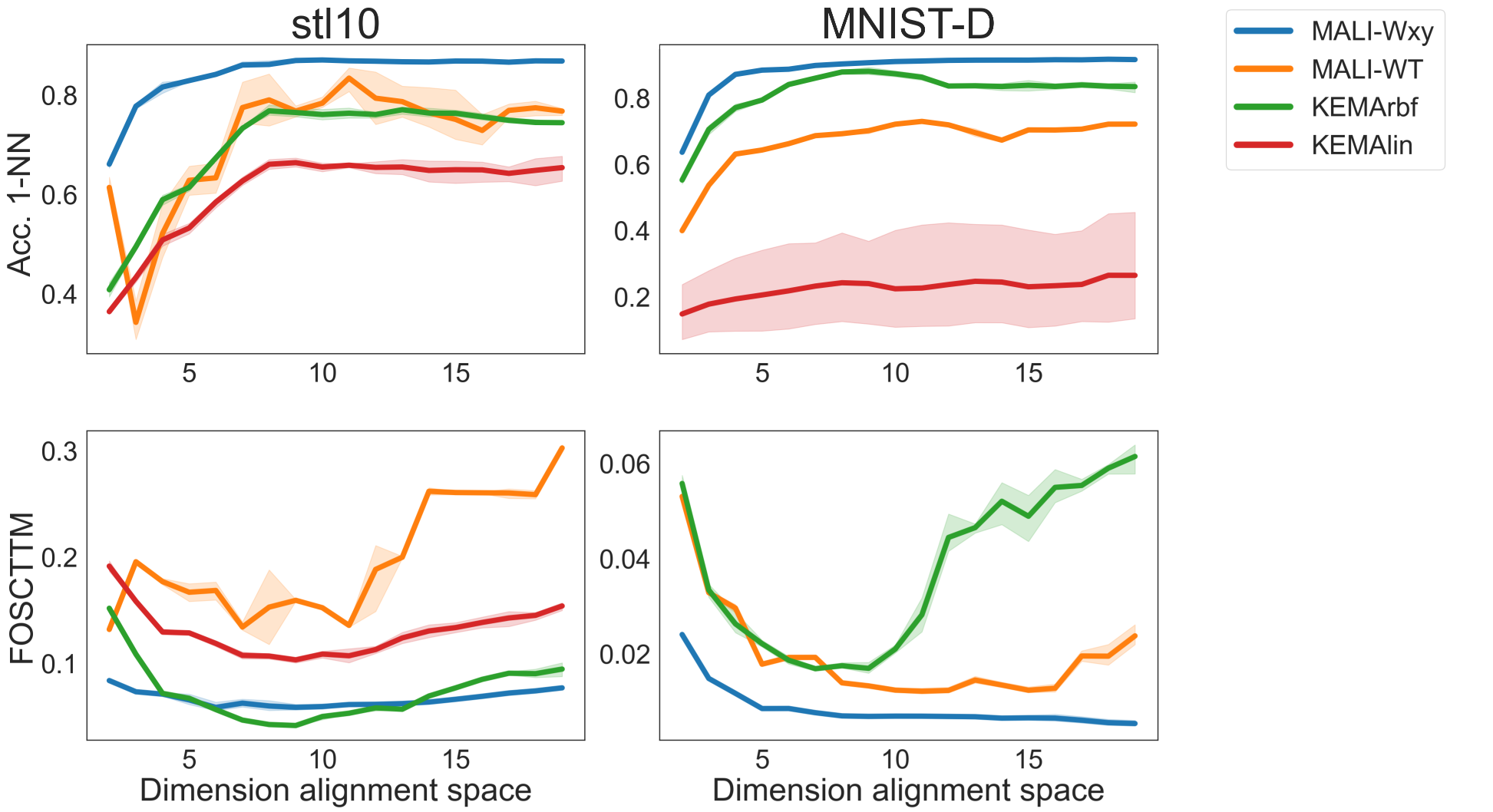}
    \caption{\textbf{Performance vs dimension of the alignment space}. The metric scores for various dimensions using the spectral embedding for each method. KEMArbf and KEMAlin present a U-shaped behavior, especially for the FOSCTTM. On the other hand MALI with $W_{\mathcal{X}\mathcal{Y}}$ is more robust to changes in the dimension of the embedding space, and consistently improves over the other approaches. KEMAlin was removed from
    the bottom right since it performs notably worse than the others and obscures the visualization.}
    \label{fig:sens_dims} 
\end{figure}

The performance of the methods is sensitive to the selected  dimensionality of the latent space obtained from the spectral embedding, especially for KEMArbf and KEMAlin. This is demonstrated in Fig.~\ref{fig:sens_dims} for both metrics computed on stl10 and MNIST-D with 50\% of the labeled data on the target domain. In this figure, the scores are computed on the learned shared latent spaces for all methods. The dimension of the alignment space ranges from 2 to 20. The experiments in Figure~\ref{fig:sens_dims} also highlight the superiority of the joint similarity from  ~\eqref{eq:w_matrix} (MALI-Wxy) in comparison to just using the $T$ matrix on the off-diagonal blocks (MALI-WT). While MALI-Wxy exhibits stable scores across dimensions, the selection of the dimension for KEMA cannot be taken lightly. Taking this into consideration, the results reported in Section~\ref{sub:metric} correspond to the dimension identified by a knee point on the plotted eigenvalues from ~\eqref{eq:KemaEig}.

\subsection{Metric performance}
\label{sub:metric}
To test the performance of the methods, it is important to analyze their behavior for various levels of labeled data in the target domain. In what follows, we include the scores for two variations of MALI. MALI-S10 is obtained when the alignment representation corresponds to the 10-dimensional spectral embedding of the joint similarities $W$, while MALI-AS is the ambient space alignment after the barycentric projection. Since the FOSCTTM metric relies on the computation of euclidean distances, MALI-AS scores are more likely to suffer from the curse of dimensionality when compared with the other methods. 

\setlength{\tabcolsep}{2pt}
\begin{table}[!h]
\caption{FOSCTTM average scores over 10 runs under various percentages of labeled data on the target domain.}
\label{tab:errors_results}
\centering
\resizebox{\columnwidth}{!}{\begin{tabular}{|lllllll|}
\hline
      & {} & \multicolumn{5}{c}{FOSCTTM} \vline \\
      &  &                                100\% &                              50\% &                              5\% &                              2\% &                              1\% \\
Dataset & Model &                                   &                                   &                                   &                                   &                                   \\
\hline
\multirow{4}{*}{Helix} & KEMAlin &                         0.298 (4) &                         0.308 (4) &                         0.251 (4) &                         0.241 (4) &                         0.242 (4) \\
      & KEMArbf &                         0.137 (3) &                         0.130 (3) &                         0.131 (3) &                         0.116 (3) &                         0.130 (3) \\
      & MALI-S10 &  {\underline{\textbf{0.033}}} (1) &  {\underline{\textbf{0.033}}} (1) &  {\underline{\textbf{0.033}}} (1) &  {\underline{\textbf{0.034}}} (1) &  {\underline{\textbf{0.033}}} (1) \\
      & MALI-AS &                \textbf{0.042} (2) &                \textbf{0.042} (2) &                \textbf{0.042} (2) &                \textbf{0.043} (2) &                \textbf{0.042} (2) \\
\cline{1-7}
\multirow{4}{*}{MNIST-D} & KEMAlin &                         0.334 (4) &                         0.333 (4) &                         0.352 (4) &                         0.378 (4) &                         0.330 (4) \\
      & KEMArbf &                \textbf{0.027} (2) &                \textbf{0.019} (2) &                \textbf{0.067} (2) &                \textbf{0.071} (2) &                \textbf{0.063} (2) \\
      & MALI-S10 &  {\underline{\textbf{0.005}}} (1) &  {\underline{\textbf{0.006}}} (1) &  {\underline{\textbf{0.018}}} (1) &  {\underline{\textbf{0.040}}} (1) &  {\underline{\textbf{0.056}}} (1) \\
      & MALI-AS &                         0.045 (3) &                         0.049 (3) &                         0.098 (3) &                         0.136 (3) &                         0.161 (3) \\
\cline{1-7}
\multirow{4}{*}{RNA-ATAC} & KEMAlin &                         0.226 (3) &                \textbf{0.159} (2) &                \textbf{0.191} (2) &                \textbf{0.182} (2) &                \textbf{0.186} (2) \\
      & KEMArbf &                \textbf{0.188} (2) &  {\underline{\textbf{0.136}}} (1) &  {\underline{\textbf{0.174}}} (1) &  {\underline{\textbf{0.173}}} (1) &  {\underline{\textbf{0.175}}} (1) \\
      & MALI-S10 &                         0.258 (4) &                         0.261 (4) &                         0.269 (4) &                         0.276 (4) &                         0.270 (4) \\
      & MALI-AS &  {\underline{\textbf{0.188}}} (1) &                         0.190 (3) &                         0.196 (3) &                         0.200 (3) &                         0.195 (3) \\
\cline{1-7}
\multirow{4}{*}{stl10} & KEMAlin &                         0.123 (4) &                         0.119 (4) &                         0.147 (4) &                         0.129 (3) &                         0.142 (3) \\
      & KEMArbf &  {\underline{\textbf{0.049}}} (1) &  {\underline{\textbf{0.056}}} (1) &                \textbf{0.087} (2) &  {\underline{\textbf{0.087}}} (1) &  {\underline{\textbf{0.096}}} (1) \\
      & MALI-S10 &                \textbf{0.054} (2) &                \textbf{0.060} (2) &  {\underline{\textbf{0.077}}} (1) &                \textbf{0.091} (2) &                \textbf{0.121} (2) \\
      & MALI-AS &                         0.117 (3) &                         0.117 (3) &                         0.147 (3) &                         0.175 (4) &                         0.195 (4) \\
\hline
\end{tabular}}
\end{table}

The FOSCTTM scores are reported in Table \ref{tab:errors_results}. Here we see that MALI-S10 greatly outperforms the KEMA methods on the Helix and MNIST-D datasets across all label percentages. For the stl10 dataset, MALI-S10  is 1st when 5\% of the data is labeled and 2nd for all other percentages, although it is not far behind KEMArbf in most of these instances.

\begin{figure}
\centering
    \includegraphics[width=\columnwidth]{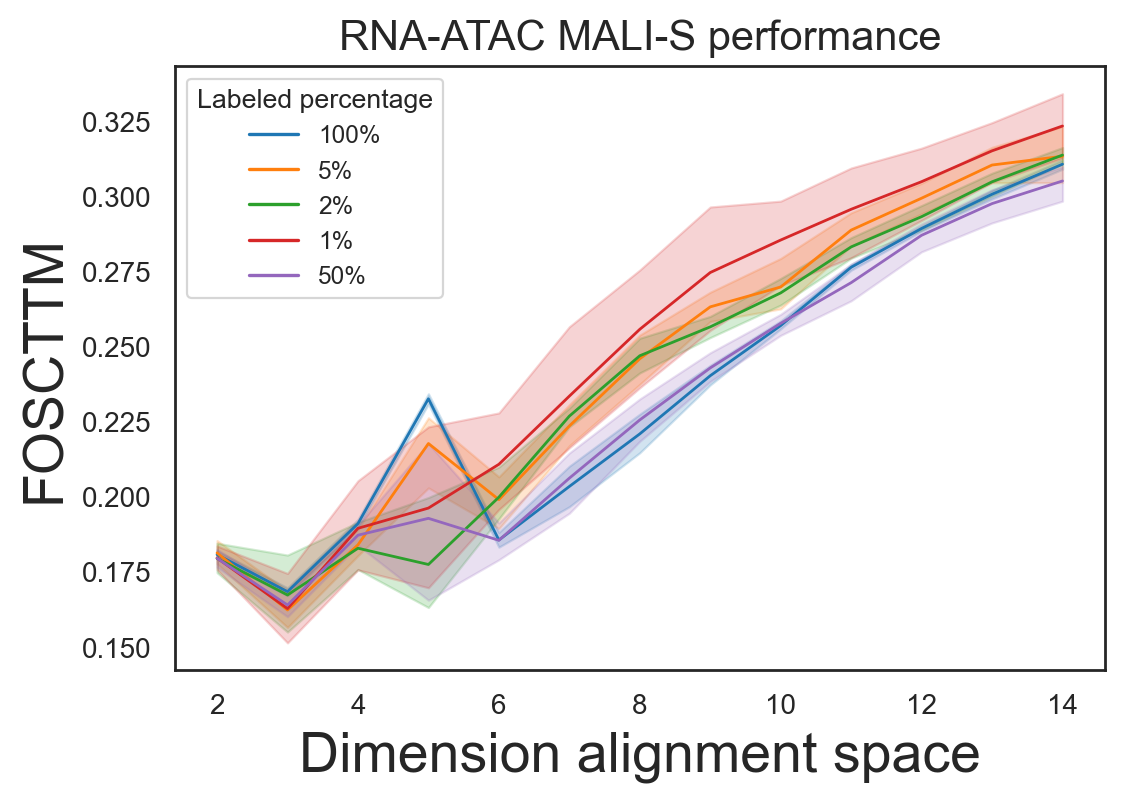}
    \caption{\textbf{MALI FOSCTTM performance vs dimension of the alignment space in the RNA-ATAC dataset}. Instead of using a 10-dimensional embedding as in the other results, we inspect the behavior of MALI on different embedding dimensions raging from 2 to 15. In this case, a smaller dimension generally results in better performance. Choosing a dimension of 3 outperforms all other methods across all label percentages except for 50\%.}
    \label{fig:mali_ed}
\end{figure}

The FOSCTTM scores for the RNA-ATAC dataset tend to favor the KEMA methods. It appears that the assumption that both domains have a similar topological structure is violated to some degree in this dataset. I.e., classes and observations that are close to each other in one domain may be far away in the other. This affects the performance of all methods (as indicated by the relatively higher error rates compared to the other datasets).  MALI is more likely to be affected by this as the DPT similarity focuses more on the global structure of the data in each domain compared to the local kernel-based similarities of the KEMA methods. However, Fig.~\ref{fig:mali_ed} demonstrates that choosing a lower dimension for the MALI spectral embedding improves the performance. In fact, choosing an embedding dimension of 3 outperforms all the other methods in Table~\ref{tab:errors_results} for all percentages except for 50\%.   

Table \ref{tab:acc_results} presents the label transfer scores. In general and in contrast with the FOSCTTM scores, we observe little discrepancy between MALI-S10 and MALI-AS. This indicates that indeed the curse of dimensionality affects MALI-AS in the FOSCTTM metric, which also explains why both MALI variants achieve closer FOSCTTM results for the 3D Helix than in the other high-dimensional datasets.  

\setlength{\tabcolsep}{2pt}
\begin{table}[!h]
\caption{Label transfer accuracy using a 1-NN classifier averaged over 10 runs under various levels of labeled data on the target domain.}
\label{tab:acc_results}
\centering
\resizebox{\columnwidth}{!}{\begin{tabular}{|lllllll|}
\hline
      & {} & \multicolumn{5}{c}{Label transfer 1-NN} \vline\\
      &  &                              100\% &                              50\% &                              5\% &                              2\% &                              1\% \\
Dataset & Model &                                   &                                   &                                   &                                   &                                   \\
\hline
\multirow{4}{*}{Helix} & KEMAlin &                         0.915 (4) &                         0.873 (4) &                         0.811 (4) &                         0.828 (4) &                         0.845 (4) \\
      & KEMArbf &  {\underline{\textbf{0.982}}} (1) &                \textbf{0.975} (2) &                         0.933 (3) &                         0.960 (3) &                         0.928 (3) \\
      & MALI-S10 &                \textbf{0.976} (2) &  {\underline{\textbf{0.976}}} (1) &  {\underline{\textbf{0.976}}} (1) &  {\underline{\textbf{0.975}}} (1) &  {\underline{\textbf{0.976}}} (1) \\
      & MALI-AS &                         0.971 (3) &                         0.971 (3) &                \textbf{0.971} (2) &                \textbf{0.972} (2) &                \textbf{0.973} (2) \\
\cline{1-7}
\multirow{4}{*}{MNIST-D} & KEMAlin &                         0.243 (4) &                         0.216 (4) &                         0.188 (4) &                         0.202 (4) &                         0.243 (4) \\
      & KEMArbf &                         0.812 (3) &                         0.835 (3) &                         0.635 (3) &                         0.632 (3) &                         0.654 (3) \\
      & MALI-S10 &                \textbf{0.918} (2) &                \textbf{0.913} (2) &  {\underline{\textbf{0.844}}} (1) &  {\underline{\textbf{0.781}}} (1) &  {\underline{\textbf{0.714}}} (1) \\
      & MALI-AS &  {\underline{\textbf{0.922}}} (1) &  {\underline{\textbf{0.920}}} (1) &                \textbf{0.836} (2) &                \textbf{0.756} (2) &                \textbf{0.682} (2) \\
\cline{1-7}
\multirow{4}{*}{RNA-ATAC} & KEMAlin &                         0.411 (4) &                         0.677 (4) &                         0.586 (4) &                         0.613 (4) &                         0.589 (4) \\
      & KEMArbf &                         0.530 (3) &                         0.702 (3) &                         0.630 (3) &                         0.623 (3) &                         0.624 (3) \\
      & MALI-S10 &                \textbf{0.755} (2) &                \textbf{0.743} (2) &                \textbf{0.734} (2) &                \textbf{0.702} (2) &  {\underline{\textbf{0.698}}} (1) \\
      & MALI-AS &  {\underline{\textbf{0.780}}} (1) &  {\underline{\textbf{0.771}}} (1) &  {\underline{\textbf{0.736}}} (1) &  {\underline{\textbf{0.711}}} (1) &                \textbf{0.695} (2) \\
\cline{1-7}
\multirow{4}{*}{stl10} & KEMAlin &                         0.571 (4) &                         0.584 (4) &                         0.546 (4) &                         0.564 (4) &                         0.539 (4) \\
      & KEMArbf &                         0.684 (3) &                         0.673 (3) &                         0.613 (3) &                         0.603 (3) &                         0.586 (3) \\
      & MALI-S10 &  {\underline{\textbf{0.879}}} (1) &  {\underline{\textbf{0.864}}} (1) &  {\underline{\textbf{0.822}}} (1) &  {\underline{\textbf{0.778}}} (1) &  {\underline{\textbf{0.717}}} (1) \\
      & MALI-AS &                \textbf{0.858} (2) &                \textbf{0.848} (2) &                \textbf{0.766} (2) &                \textbf{0.690} (2) &                \textbf{0.636} (2) \\
\hline
\end{tabular}}

\end{table}

\begin{figure}[!h]
    \includegraphics[width = \columnwidth]{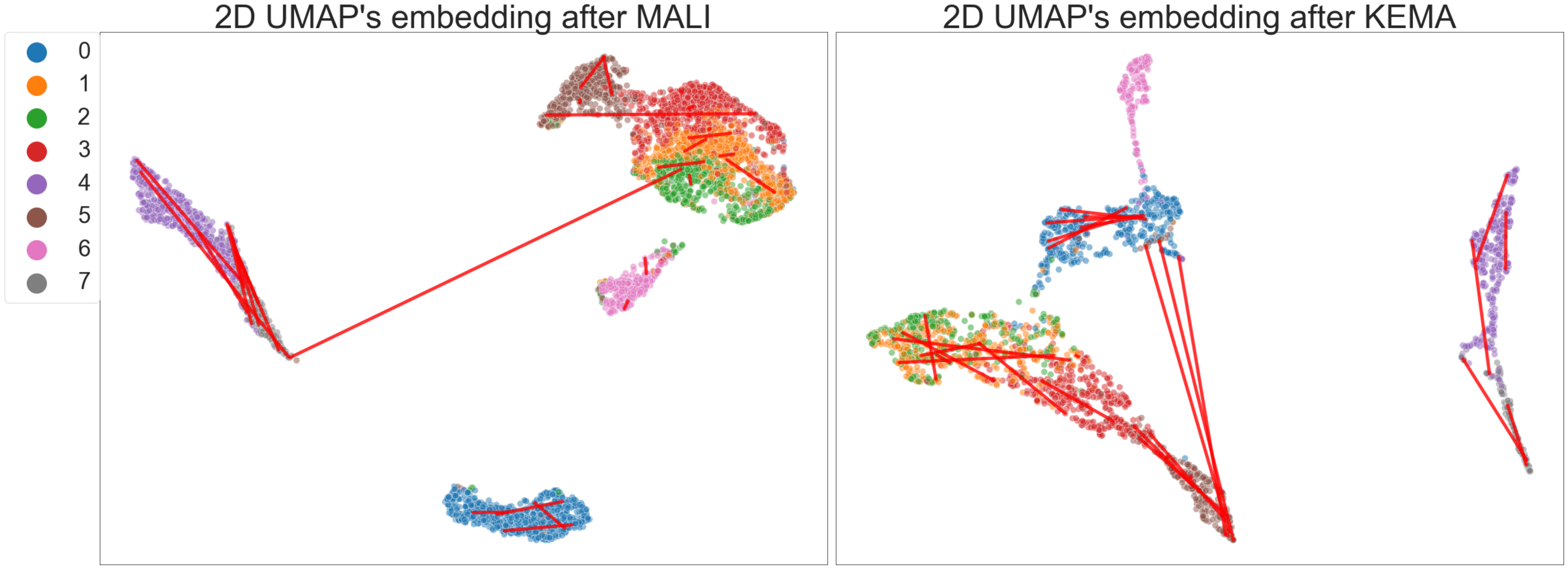}
    \caption{\textbf{UMAP embeddings of RNA-ATAC data after alignment}. Embeddings obtained by applying UMAP~\cite{mcinnes2018umap} with a precomputed distance matrix $1-W$ comparing  MALI and KEMA alignments.  Red lines connect the ground truth paired data points for a random subset of the data. Shorter lines are indicative of a better alignment. Even though both methods obtain similar FOSCTTM scores, MALI maintains a cleaner class separation, especially for cells with labels 1 and 2. This is reflected in the accuracy scores of Table~\ref{tab:acc_results}.}
    \label{fig:2d_rna_atac}
\end{figure}

\begin{figure}[!h]
    \centering
    \includegraphics[width = \columnwidth]{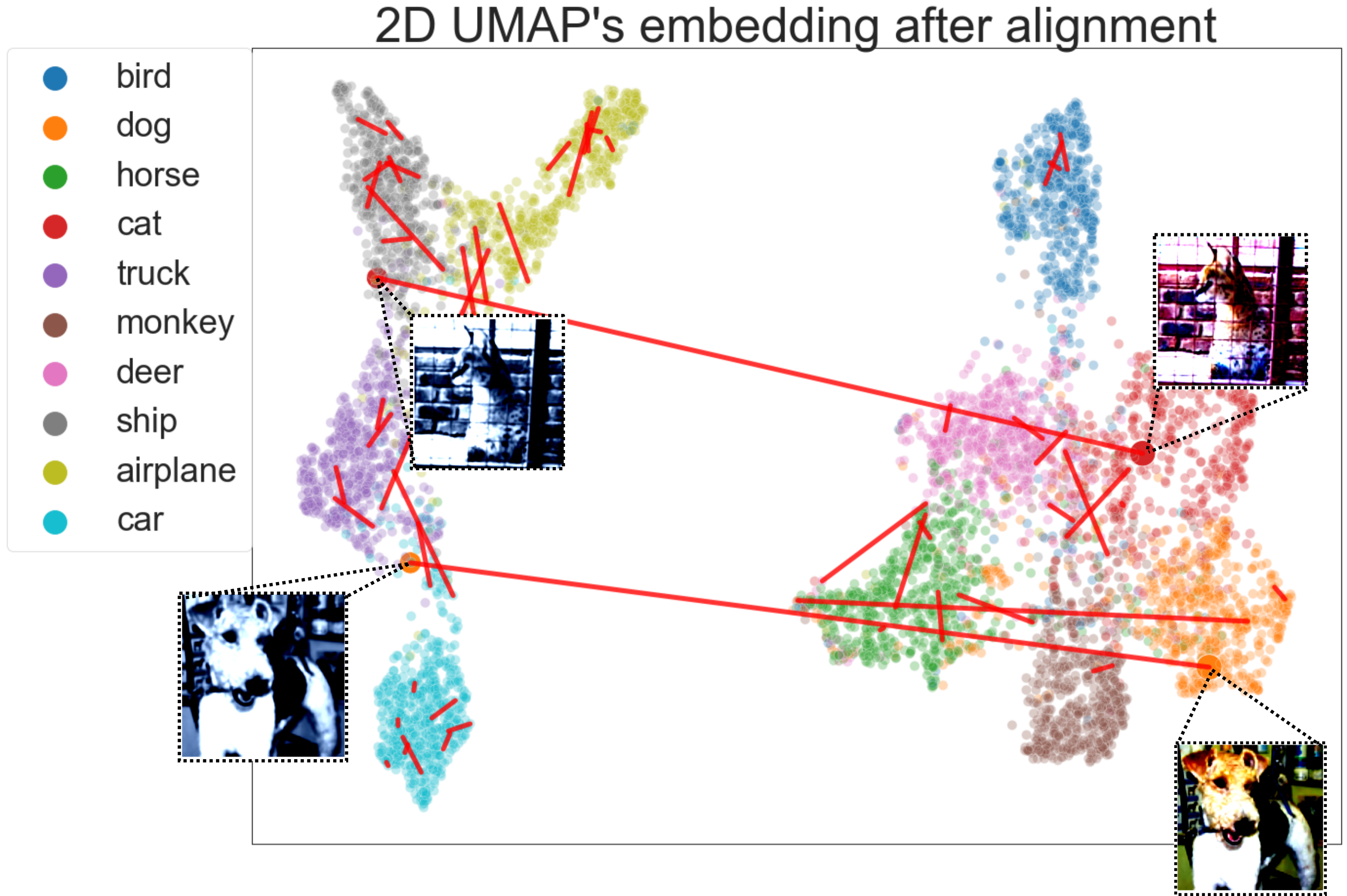}
    \caption{\textbf{UMAP embedding of stl10  after MALI alignment}.  Samples are colored by their class label and red lines connect the ground truth paired images for a subset of the data. Shorter lines are indicative of a better alignment. We display the two images with the worst alignment score. Most points are correctly aligned within their class, leading to a relatively high classification accuracy in the domain adaptation problem.}
    \label{fig:2d_mali_stl10}
\end{figure}

While KEMArbf is competitive with MALI in the FOSCTTM metric, its performance in label transfer is largely inferior. The main reason is that KEMA projections sometimes cause classes to overlap to a higher degree than MALI. This is clear in Fig.~\ref{fig:2d_rna_atac} in which classes 1 and 2 are distinguishable in the MALI UMAP embedding, but are greatly overlapped in the KEMA embedding. Figure~\ref{fig:2d_mali_stl10} also shows that MALI  does well at maintaining class separability in the stl10 dataset. Thus, the 1-NN classifier attains a much higher accuracy in these datasets compared to the KEMA methods. 

\subsection{Soft assignments with entropic regularized OT}
\label{sub:soft}
MALI is not restricted to producing one-to-one matchings as we obtained in the previous section. In many cases, we might be interested in finding a soft matching between domains. For such a task, imposing an entropic regularization by setting $\epsilon > 0$, is a natural alternative. It forces the transport plan $T$ to find a dense solution instead of a sparse one. Thus, we can interpret the learned entries of $T$ as a soft matching between samples in both domains.

\begin{figure}[!h]
    \includegraphics[width = \columnwidth]{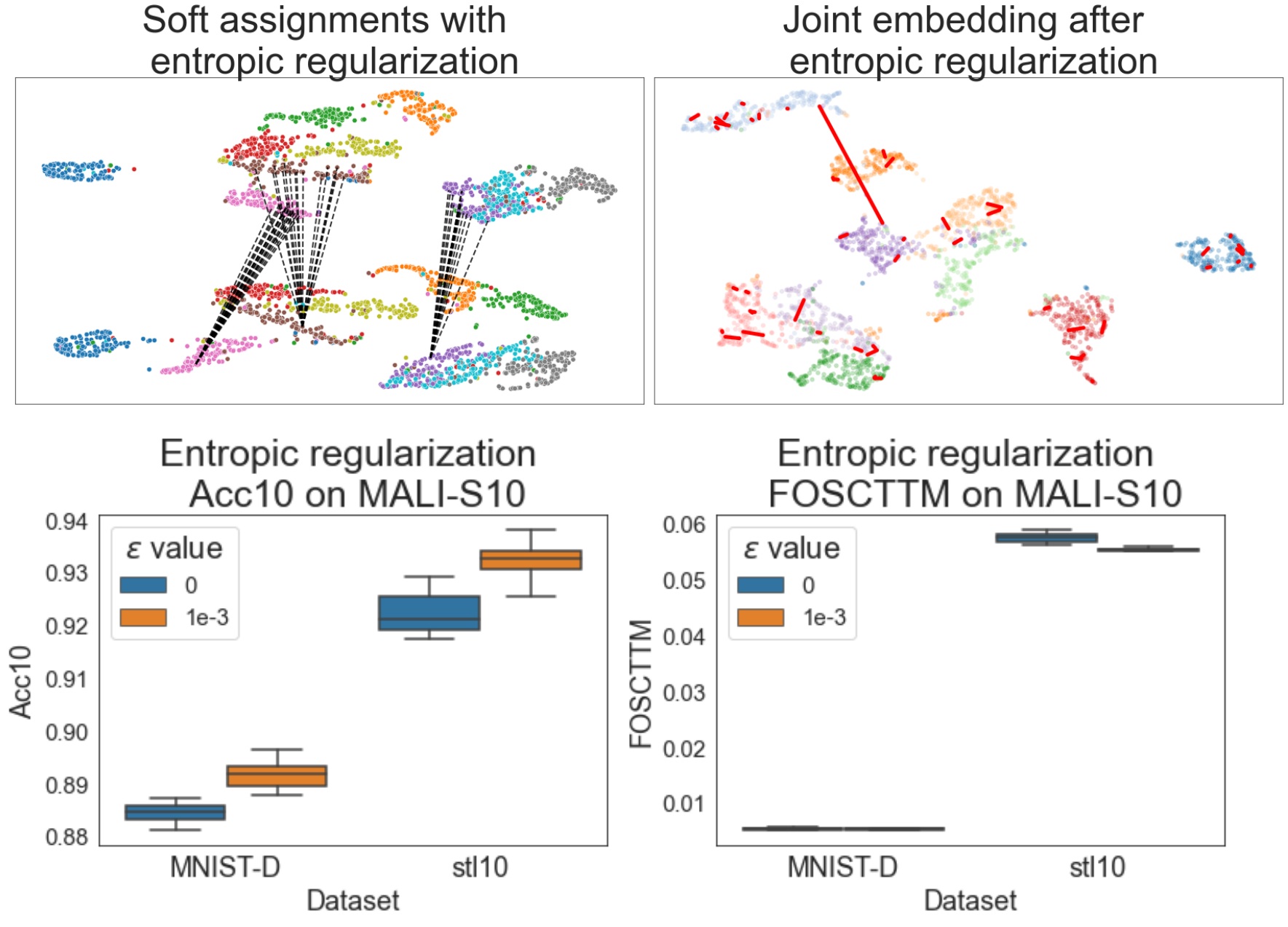}
    \caption{\textbf{Soft matchings with entropic regularization}. The \textbf{top-left} subplot shows the assignments from individual points in $\mathcal{X}$ to $\mathcal{Y}$ in the MNIST-D dataset. Instead of matching in a one-to-one fashion, each point finds a collection of
    matches with high similarity. \textbf{Top-right}, we computed a UMAP embedding using the joint similarity matrix $W$ built with a dense matrix $T$. To represent the goodness of alignment, we highlight randomly selected red lines connecting ground truth matches. Most of the connections are short, with longer connections being an artifact of the UMAP algorithm. Overall, the alignment not only looks qualitatively accurate, but as displayed in the bottom panels the performance metrics  score better with $\epsilon = 0.001$ than for the unregularized case ($\epsilon = 0$). Acc10 is the label transfer accuracy using a 10-NN classifier.}
    \label{fig:entropy_reg}
\end{figure}

Figure \ref{fig:entropy_reg} exemplifies how including the entropic regularization affects the solutions. Here we show on the MNIST-D dataset how each point can be matched with multiple points with high similarity. Using the dense $T$ matrix to construct the joint similarity matrix $W$ results in a good embedding with relatively few large errors. This is corroborated by our metrics which appear to show an improvement when including the regularization. This is likely because the soft assignments created with the entropy regularization are more robust to local changes in the neighborhood of a given data point.  

\subsection{Unbalanced number of observations}
\label{sub:unbalance}
\begin{figure}[!h]
\centering
    \includegraphics[width =  \columnwidth]{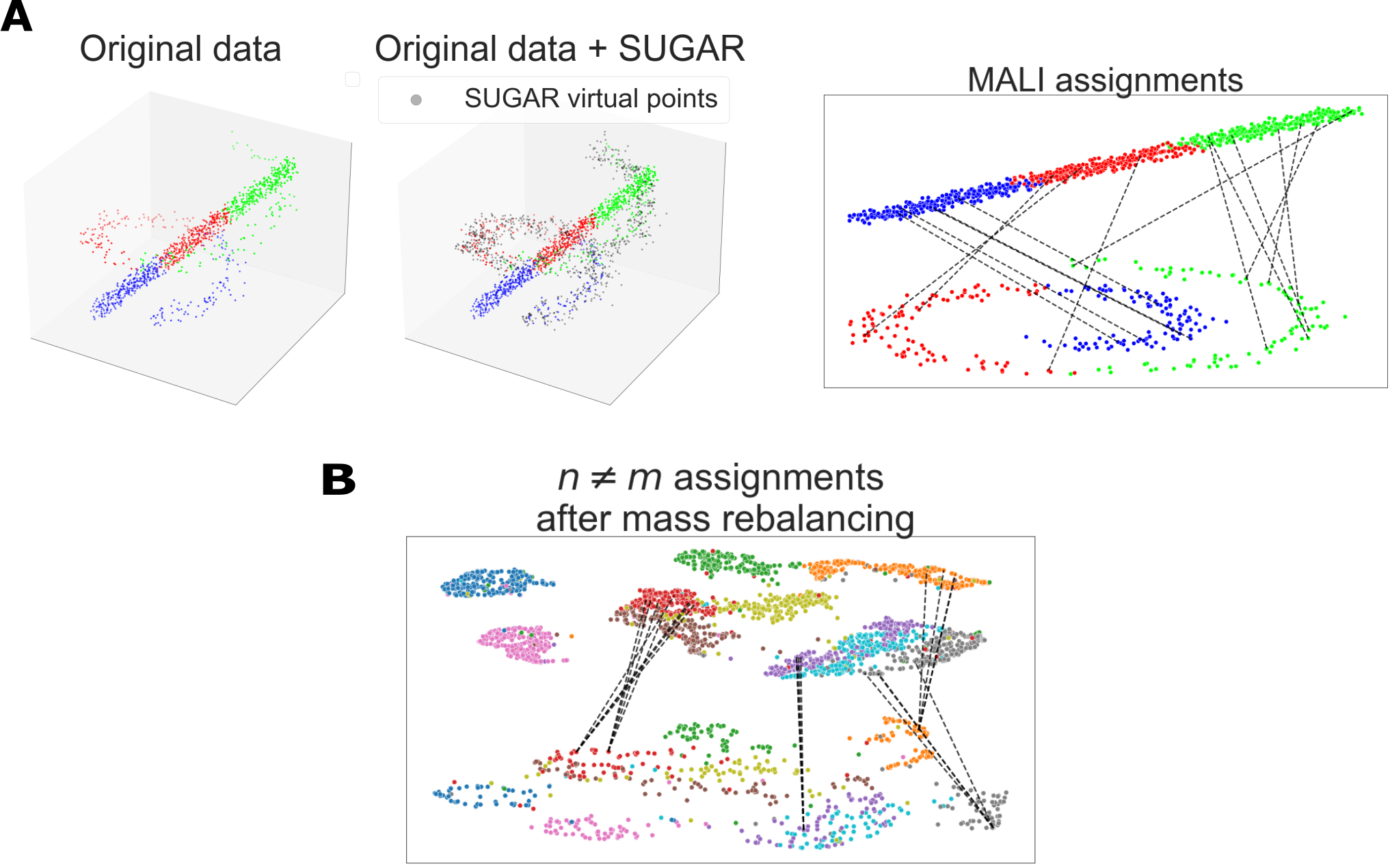}
    \caption{\textbf{A)} Oversampling and density equalization via SUGAR. This strategy allows us to avoid rebalancing the masses, since we can synthetically reproduce the $n = m$ case. \textbf{B)} Assignments for a low-sampled $\mathcal{X}$ domain after rebalancing the masses. Each sample from $\mathcal{X}$ is assigned to multiple counterparts in $\mathcal{Y}$ due to their higher mass.}
    \label{fig:sugar}
\end{figure}

So far, we have assumed that we have the same number of observations for both domains, i.e., $n = m$. The extension to the  case $n \neq m$ can be handled in many ways. Here, we present two relatively straight-forward solutions. One is to rebalance the masses. Without loss of generality we can set $b = \frac{a \times n}{m}$, where $b$ and $a$ are the individual masses assigned uniformly to all samples in domains $\mathcal{Y}$ and $\mathcal{X}$, respectively. This enables a soft assignment for some or all of the observations in either of the domains. Figure~\ref{fig:sugar} shows an example of this.
 
We note that the masses can also be modified when $n = m$, resulting in soft assignments for some or all the observations. The specific problem will determine the best approach for modifying the masses.  For instance, if the data density is lower for a given data region in one domain compared to its counterpart in the other domain, one may consider increasing the masses of a set of samples belonging to the low density region. 

Another simple but powerful solution to the imblanced problem is to oversample with density equalization via a method such as SUGAR \cite{lindenbaum2018geometry}. In this case, both domains will be balanced with the same amount of samples and uniform densities and the problem reduces to the $n = m$ case. The generated points can then be eliminated after performing the alignment and embedding. Figure~\ref{fig:sugar} shows an example of this.

\subsection{Kernel hyperparameters}

\begin{figure}[!h]
\centering
    \includegraphics[width=\columnwidth]{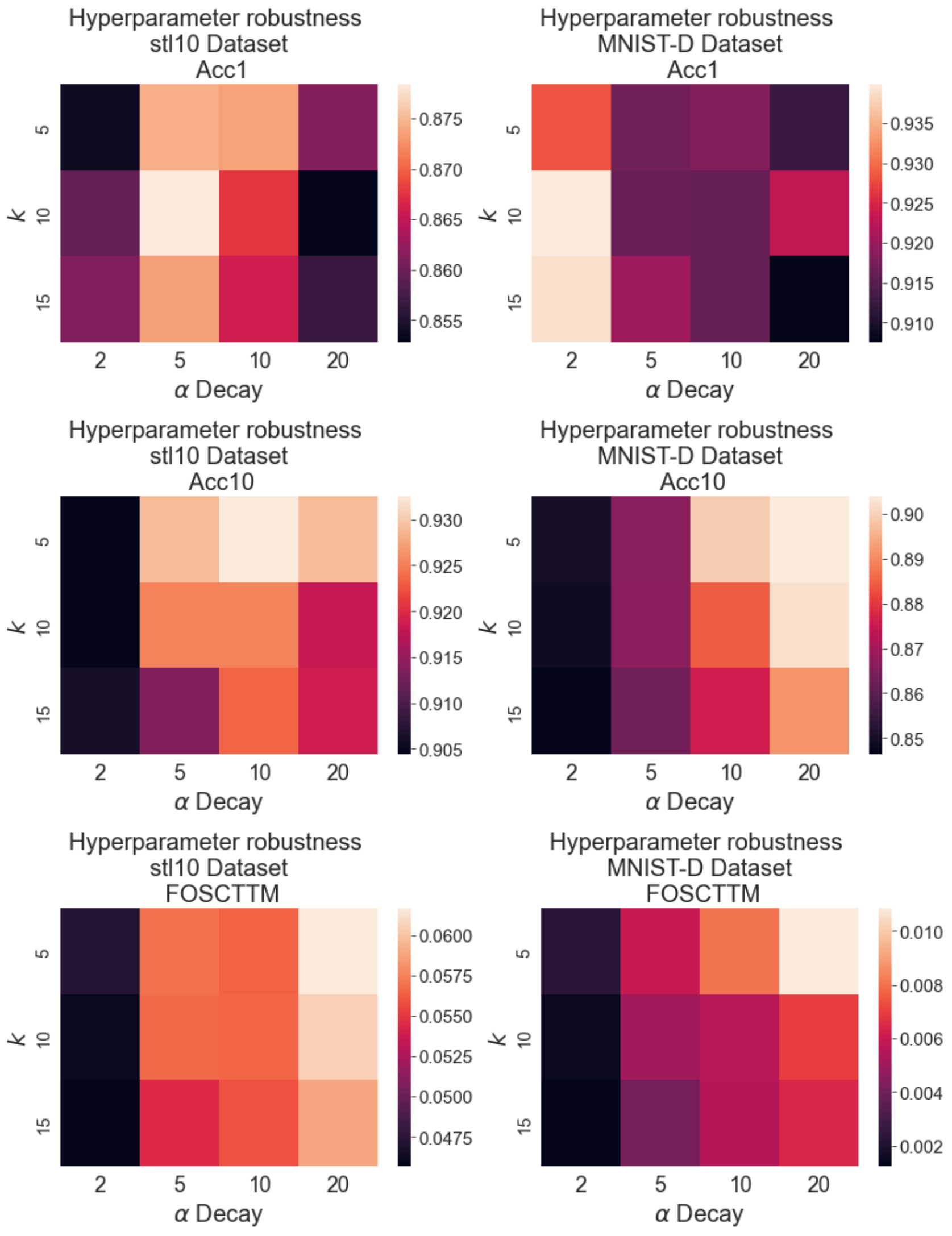}
    \caption{\textbf{Kernel hyperparameters robustness}. For different hyperparameter choices, we computed the performance metrics for stl10 and MNIST-D, and report their average value after 15 runs. In general, there is not a significant drop in performance for common choices of the parameters (note the scales for each heatmap). This shows that when using the $\alpha$-decay kernel, MALI is robust to hyperparameter selection. Acc1 and Acc10 are the 1- and 10-NN classifier accuracies, respectively.}
    \label{fig:hyperparams}
\end{figure}

 MALI does not depend on our particular kernel choice, which is the $\alpha$-decay kernel shown in (\ref{eq:DecayGaussKern}). Any kernel that captures a sensible geometry in both domains is well suited for the task. Nevertheless, in Figure \ref{fig:hyperparams} we include a sensitivity analysis for the particular hyperparameters employed in this paper, $\alpha$ and $k$. Overall, we did not find a compromising drop in performance for any of the metrics when we vary both hyperparameters between common value choices. In many cases there is an improvement over the reported results in Section \ref{sec:results}, for which we set the values $\alpha = 10$ and $k = 10$.

\section{Conclusion}

We presented MALI, a manifold alignment method capable of finding a common meaningful representation for two distinct but related domains. MALI only requires side coarse information to perform the alignment, such as discrete labels in both domains. MALI combines the diffusion geometry of the data alongside the labels to find inter-domain distances, which are then used to couple the datasets via optimal transport. This coupling can be used to obtain a shared representation of both domain. Our method improves over other related manifold alignment methods designed for this setting, especially in the domain adaptation problem.

\bibliography{ref}
\bibliographystyle{siam}

\end{document}